\documentclass[lettersize,journal]{IEEEtran}
\usepackage{framed,multirow}
\usepackage{tabularx}
\usepackage{booktabs}
\usepackage{amsmath,amsfonts}
\usepackage{algorithmic}
\usepackage{algorithm}
\usepackage{array}
\usepackage[caption=false,font=footnotesize,labelfont=rm,textfont=rm]{subfig}
\usepackage{textcomp}
\usepackage{stfloats}
\usepackage{url}
\usepackage{verbatim}
\usepackage{graphicx}
\usepackage{cite}
\usepackage[T1]{fontenc}
\usepackage{booktabs}
\usepackage{multirow}
\usepackage{footnote}
\usepackage{threeparttable}
\usepackage{svg}

\begin{document}

\title{Motion-Guided Small MAV Detection in Complex and Non-Planar Scenes}

\author{Hanqing Guo, Canlun Zheng, Shiyu Zhao\vspace{-2em}
% <-this % stops a space
\thanks{H. Guo is with the College of Computer Science and Technology, Zhejiang University and also with the School of Engineering, Westlake University, Hangzhou, China (e-mail: guohanqing@westlake.edu.cn)}
% <-this % stops a space
\thanks{C. Zheng and S. Zhao is with the School of Engineering at Westlake University, Hangzhou, China (e-mail: zhaoshiyu@westlake.edu.cn)}

\thanks{(\textit{Corresponding author: Shiyu Zhao.})}
}

%\markboth{IEEE Transactions on xxx,~Vol.~14, No.~8, MAY~2023}%

\maketitle

\begin{abstract}
In recent years, there has been a growing interest in the visual detection of micro aerial vehicles (MAVs) due to its importance in numerous applications. However, the existing methods based on either appearance or motion features encounter difficulties when the background is complex or the MAV is too small. In this paper, we propose a novel motion-guided MAV detector that can accurately identify small MAVs in complex and non-planar scenes. This detector first exploits a motion feature enhancement module to capture the motion features of small MAVs. Then it uses multi-object tracking and trajectory filtering to eliminate false positives caused by motion parallax. Finally, an appearance-based classifier and an appearance-based detector that operates on the cropped regions are used to achieve precise detection results. Our proposed method can effectively and efficiently detect extremely small MAVs from dynamic and complex backgrounds because it aggregates pixel-level motion features and eliminates false positives based on the motion and appearance features of MAVs. Experiments on the ARD-MAV dataset demonstrate that the proposed method could achieve high performance in small MAV detection under challenging conditions and outperform other state-of-the-art methods across various metrics.
\end{abstract}


\section{Introduction}
Vision-based air-to-air MAV detection has attracted increasing attention in recent years due to its application in many tasks such as vision-based swarming\cite{tang2018vision}, aerial see-and-avoid\cite{park2023vision}, and malicious MAV detection\cite{2022detection}. Air-to-air MAV detection is \textit{more challenging} than general object detection because the camera is moving and the target MAV is often engulfed by complex background scenes such as buildings and trees when the camera looks down from the top. Moreover, the target MAV may be extremely small in the image when seen from a distance.

\begin{figure*}[t]
\centering
	\subfloat[Complex background.]{\includegraphics[width=0.48\linewidth]{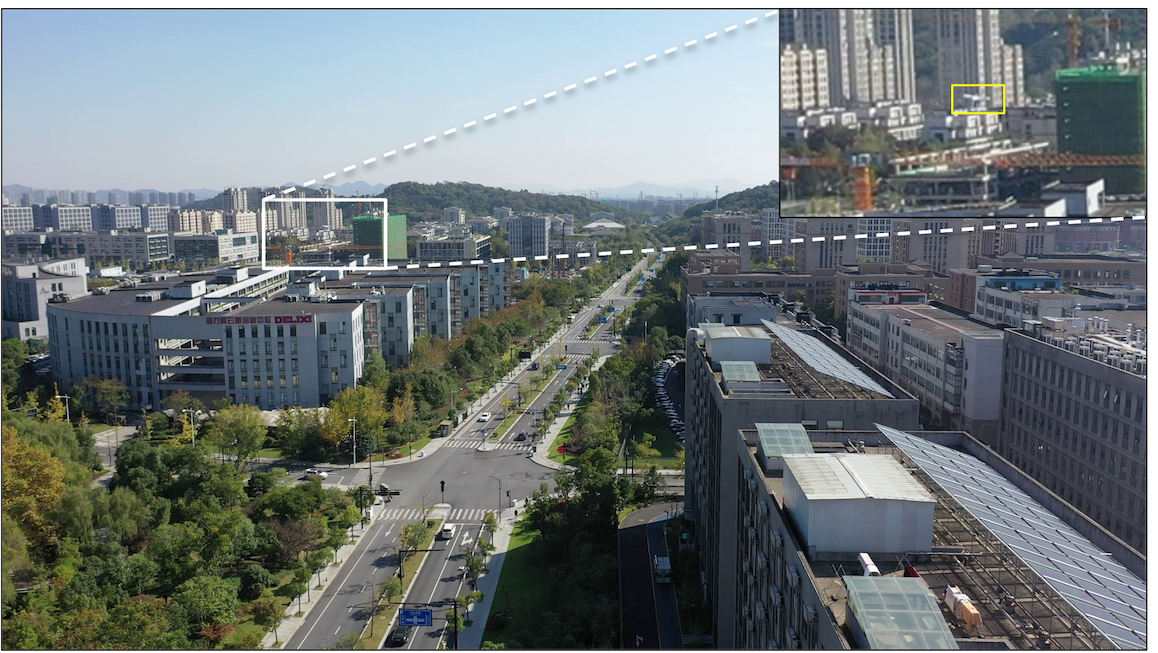}}\hspace{0.5mm}
    \subfloat[Small MAV targets (8x8 pixels).] {\includegraphics[width=0.48\linewidth]{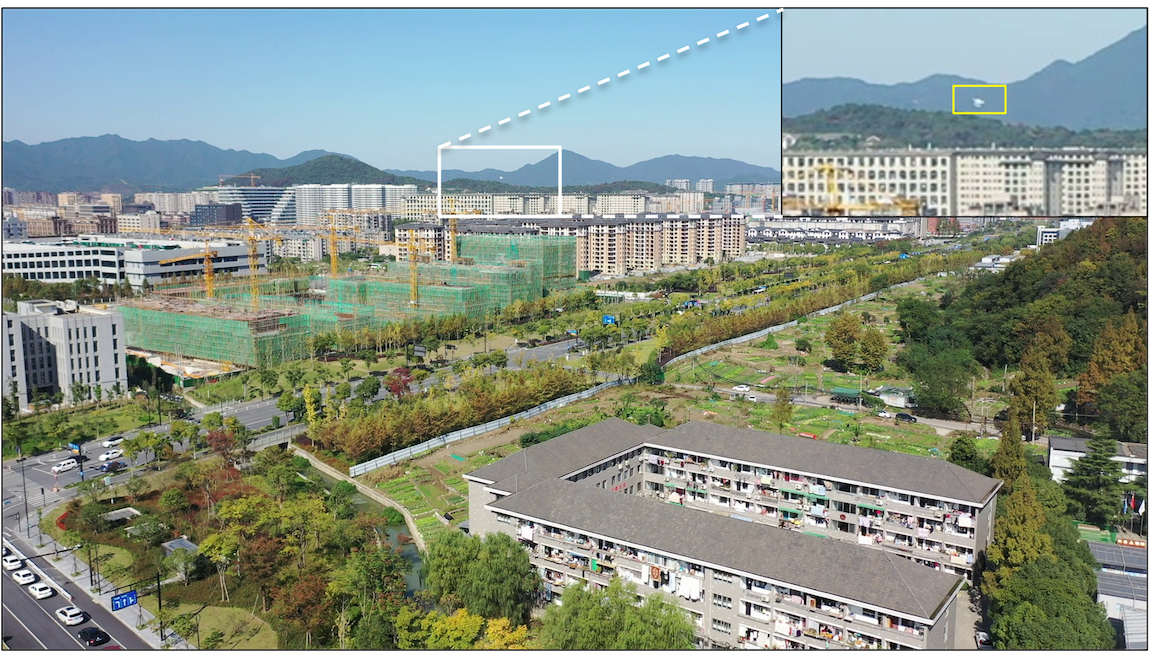}}
	\caption{Examples of challenging conditions for MAV detection. In air-to-air MAV detection scenarios, the target MAV is often engulfed by a complex background or may be extremely small in the image. The yellow boxes
enclose MAVs. (Top right images are a better view with 300$\%$ zoom-in).}
	\label{fig_1}
\end{figure*}

Recently, many appearance-based methods that rely on deep learning techniques have been proposed for vision-based MAV detection. For example, some state-of-the-art object detection networks such as YOLO series, R-CNN series, SSD, and DETR have been applied to MAV detection \cite{2021DT-Benchmark, 2021Air, 2022Anti-UAV-DT}. These methods can work effectively in relatively simple scenarios where the target MAV is distinct from the background and its size is relatively large in the image. However, they often fail in more complex scenarios since appearance features are \textit{not reliable}. For example, as shown in Fig.~\ref{fig_1}(a), when the background scene is extremely complex, the target MAV can easily get engulfed in the background scenes. Moreover, when the target MAV flies far away from the camera, its image may only occupy a tiny portion of the image. For example, as shown in Fig.~\ref{fig_1}(b), an MAV seen from about 100~m away only occupies 8$\times$8 pixels in an image of 1920$\times$1080 pixels. Therefore, high-performance algorithms must be developed to detect MAVs under such challenging conditions.

\begin{figure}[h]
	\centering
	\subfloat[The RAFT \textit{succeeds} in extracting motion features when the target is large or the background is simple.]{\includegraphics[width=0.485\linewidth]{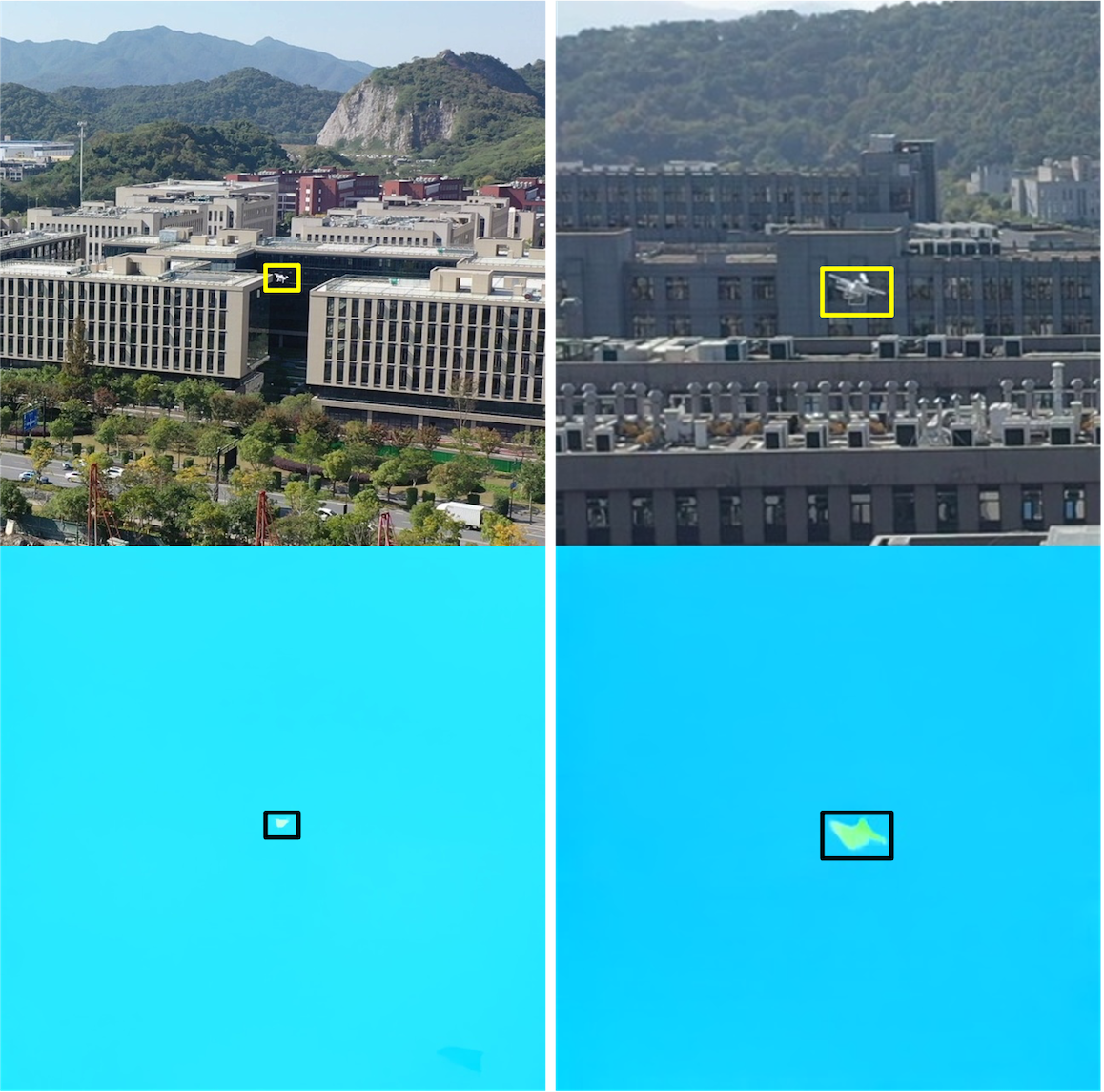}}\hspace{1mm}
	\subfloat[The RAFT \textit{fails} to extract motion features when the target is small or the background is complex.]{\includegraphics[width=0.485\linewidth]{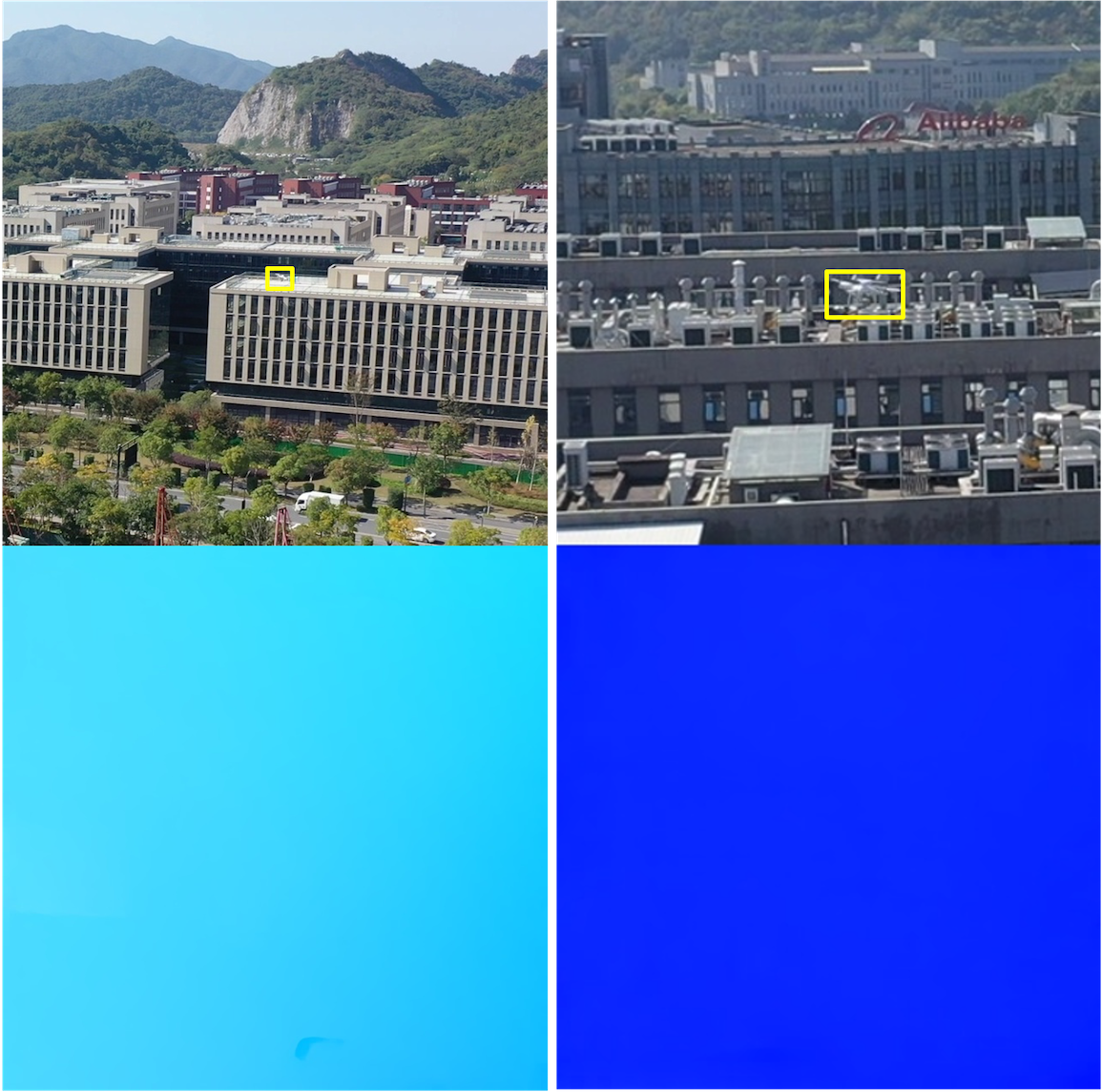}}
	\caption{Testing results of the dense optical flow of moving MAVs extracted by the pre-trained model of the RAFT \cite{2020RAFT}. \textbf{Yellow box} indicates the MAV target. \textbf{Black box} indicates the extracted dense optical flow. }
	\label{fig2}
\end{figure}

Motion features can greatly help detect MAVs under challenging conditions. Many motion-assisted methods introduce motion features into MAV detection based on, for example, background subtraction\cite{park2023vision, 2021Fast}, low-rank based methods\cite{2017UDT, wang2019flying}, spatio-temporal information\cite{2022Camera, Xie2021SmallLT}, and optical flow \cite{2021Dogfight, wang2023RAFT}. However, motion-assisted MAV detection still faces the following \textit{challenges}. First, motion cues of MAVs are difficult to separate from the background when the target size is too small or the background is too complex. For example, we tested the popular optical flow model RAFT \cite{2020RAFT} for motion feature extraction. The result is given in Fig.~\ref{fig2}, which shows that the dense optical flow network can successfully extract the motion information when the target has a high contrast to the background or is large enough. However, when the background is complex or the target is small, the motion information can hardly be extracted. Second, the influence of image alignment error or parallax effects is significant when the camera is moving. Most of the existing motion-assisted methods only deal with \textit{planar} backgrounds \cite{park2023vision, 2021Fast, zhang2021jointly} so that affine transformation or 2D perspective transformation can be used to align adjacent frames. Although the planar assumption is valid when the camera flies at a high altitude so that the height of the ground objects is negligible, it is still invalid in many cases when the flight altitude is relatively low. Objects on the ground such as buildings, trees, and lampposts may violate the planar assumption and result in many false positives.

To overcome the limitations of the existing approaches, we propose a motion-guided MAV detector (MGMD) for small MAV detection in complex and non-planar scenes. This algorithm consists of three modules: motion feature enhancement, trajectory filtering, and local fine detection. First, a motion feature enhancement module is applied to extract the motion features of small MAVs from the complex background. Second, multi-object tracking is used to generate trajectories for each target and a trajectory-based classifier is utilized to eliminate false positives. Third, an appearance-based classifier and an appearance-based detector that operates on the cropped regions are used to achieve precise detection results.

The technical novelties are summarized as follows.

\begin{itemize}
\item[\footnotesize\textbullet] Compared with traditional background subtraction and optical flow, our proposed motion feature enhancement module can effectively extract motion features of extremely small MAVs from dynamic and complex backgrounds. This is because we aggregate pixel-level motion features by frame alignment and multi-frame differences. Our proposed detector outperforms the state-of-the-art algorithms such as \cite{2021Dogfight, TPH-YOLOv5, mega} on various evaluation metrics.

\item[\footnotesize\textbullet] The influence of false positives generated by motion parallax can be effectively removed by multi-object tracking, trajectory filtering, and appearance-based classification. This is because our proposed method can model the spatial and temporal features of MAVs that are different from the false positives generated by motion parallax. The effectiveness of this method is validated by experimental results on the ARD-MAV dataset where 3D structures such as tall buildings, trees, and lampposts are dominant.
\end{itemize}

\section{Related work}
\subsection{Vision-based MAV detection}
Vision-based MAV detection has been a hot topic in recent years. Due to the great success on general object detection, some researchers have directly applied the state-of-the-art object detection networks for MAV detection. The work in \cite{2021Air} evaluates eight state-of-the-art deep learning algorithms on the Det-Fly dataset for MAV detection. Similarly, the authors in \cite{2021DT-Benchmark} evaluate four state-of-the-art deep learning algorithms on three representative MAV datasets. To further improve the detection accuracy, the authors in \cite{2021_Transfer_Adaptive_fusion} propose a novel approach that combines transfer learning and adaptive fusion mechanism to improve small object detection performance. However, the experimental results in these works show that the existing object detection networks do not perform well in challenging conditions such as complex backgrounds, motion blur, and small objects due to insufficient visual features.

To improve the detection accuracy under challenging conditions, the motion feature is introduced to assist MAV detection. \cite{Rozantsev2017DetectingFO} is an early work in the field of detecting flying objects using a moving camera. The authors first employ two CNN networks in a sliding window fashion to obtain the motion-stabilized spatial-temporal cubes and then use a third CNN network to classify MAV in each spatial-temporal cube. Similarly, the work in \cite{2021Fast} proposes a UAV-to-UAV video dataset and a background-subtraction-based method for small UAV detection. Subsequently, \cite{2021Dogfight} proposes a two-stage segmentation-based approach for detecting drones from drone videos and \cite{wang2023RAFT} proposes a feature super-resolution-based UAV detector with motion information extracted by dense optical flow. Recently, the work in \cite{2022transvisdrone} proposes an end-to-end method for drone-to-drone detection. The authors utilize CSPDarkNet53 to learn spatial features and VideoSwin Transformer to exploit temporal information. These methods could work fine when targets occupy a relatively large part of the entire image or the background is comparatively simple. However, when handling extremely small targets (such as 10$\times$10 pixels) in a cluttered environment, the detection accuracy degrades vastly.

\subsection{Small object detection in aerial images}
Object detection in aerial images has been a popular task in recent years. Due to the extremely small object sizes and numerous background clutters, the general object detection algorithms usually perform poorly in such scenarios. \cite{TPH-YOLOv5} tries to improve the object detection accuracy on drone-captured scenarios by adding the Transformer Prediction Heads and CBAM to YOLOv5. The work in \cite{PRL_Dense_Similar} proposes a dense-and-similar object detector with four key components: a coarse detector, an adaptive clustering procedure, a similar-class classifier, and a fine detector. Recently, the work in \cite{PRL_Small_UAV_Scenes} aggregates abundant object-oriented information in low-resolution imagery to enhance the recognition ability of small objects in large-scale images. In addition, temporal context has also been introduced to enhance the feature of small targets. The work in \cite{2018clusternet} proposes a novel two-stage spatio-temporal CNN which effectively and efficiently combines both appearance and motion information. The work in \cite{2023temporal} presents a spatiotemporal deep learning model based on YOLOv5 that exploits temporal context by processing sequences of frames at once. Although these methods are shown to be effective in certain small object detection scenarios, their effectiveness for small MAV detection has not been verified yet. Moreover, the computational cost of them is usually unacceptable for real-world applications. 

\begin{figure*}[!t]
	\centering
	\includegraphics[width=0.99\linewidth]{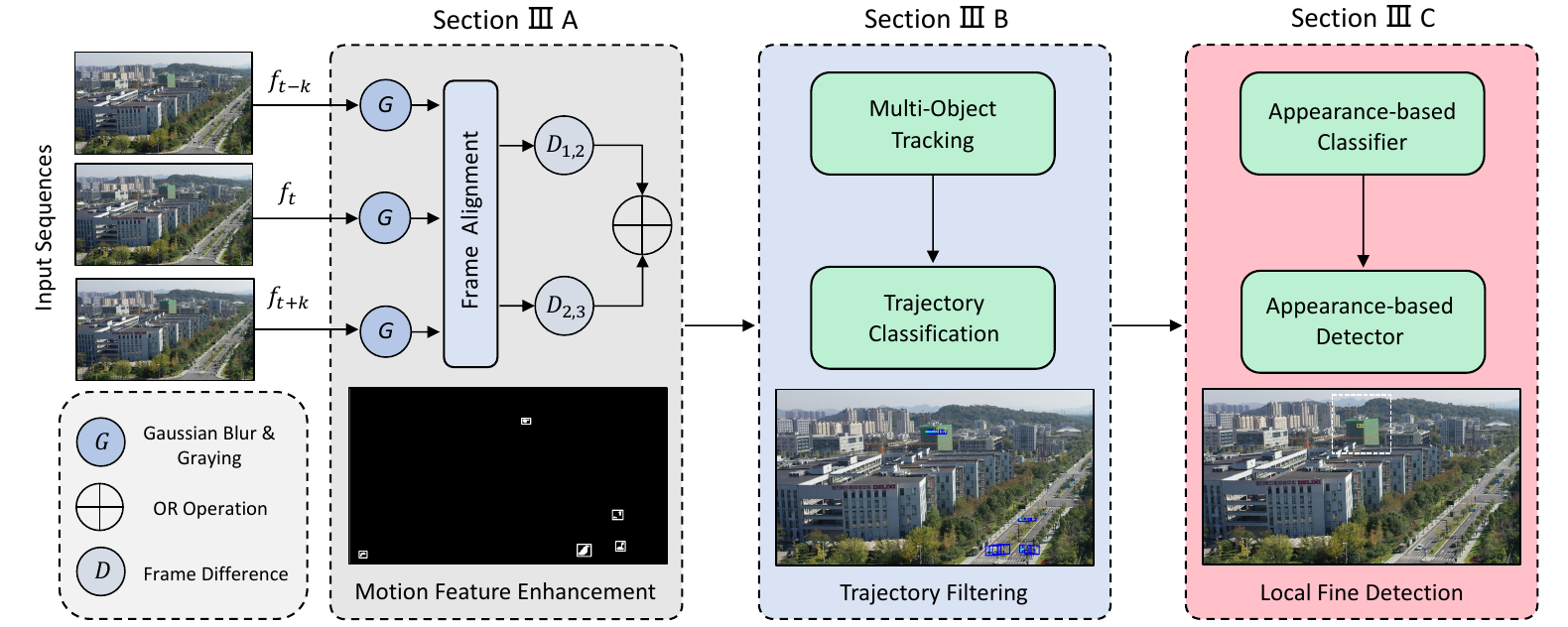}
	\caption{The architecture of the proposed MGMD algorithm. First, we use frame alignment and three-frame difference to segment moving objects from complex backgrounds. Then, multi-object tracking is applied to generate trajectories for each detection, and a trajectory-based filter is used to eliminate false positives. Finally, a local appearance-based classifier and a local appearance-based detector are utilized on a cropped region to obtain precise detection results.}
	\label{fig3}
\end{figure*}

\section{Proposed method}\label{methodology}
The architecture of our proposed algorithm is shown in Fig.~\ref{fig3}. It is composed of three modules: motion feature enhancement (MFE), trajectory filtering (TF), and local fine detection. The following sections detail each module.

\subsection{Motion feature enhancement} Since the MAV is difficult to detect when its size is too small or the background is too complex, we first design a motion feature enhancement module to obtain the potential areas that probably contain moving MAVs.

Motion feature extraction or moving object detection is a well-studied problem in constrained environments where the camera's ego motion is not obvious. However, when it comes to unconstrained ego motion, moving object detection becomes a much harder problem. Inspired by related works \cite{2021Fast, 2018effective, deepfusion2021}, we first exploit image alignment techniques to obtain stabilized frames, and then frame difference and morphological operations are performed to segment moving objects from multiple frames. Our motion feature enhancement module can be divided into three steps: frame alignment, motion feature extraction, and post-processing.

\subsubsection{Frame alignment.}
To separate moving objects from the dynamic background, we must first align successive frames so that the influence of the camera's ego motion can be eliminated. In this paper, 2D perspective transformation is used for motion compensation because it exactly models the 2D background motion when the background results from the relative motion of a 2D plane in the 3D world. In principle, the perspective transformation requires the background to be planar or the camera only rotates. In some cases, this is a reasonable approximation since the MAV looks at the ground from a high altitude. However, when the flight altitude is relatively low, 3D objects such as tall buildings, lampposts, trees, and wire poles can remarkably influence the motion compensation quality and generate many false positives. Nevertheless, our proposed trajectory-based filter and the appearance-based classifier are applied to remove these false positives. 

For computational efficiency and robustness in textureless regions such as sky and grassland, grid-based key points are used instead of the feature detectors to calculate the homography matrix. We sample $30 \times 20$ key points uniformly distributed in each row and column across the previous $k$ frame. Then, these key points are tracked by the pyramidal Lucas-Kanade (LK) algorithm to obtain the corresponding points in the current frame. After these key points are matched over two successive frames, the homography matrix $H$ is calculated with the RANSAC method to reject outliers. The image in the previous $k$ frame $I_{t-k}$ can be aligned with the current frame $I_t$ by the perspective transformation. %$\hat {I}_{t-k}= \emph{H} I_{t-k}$\cite{PLK}.
\begin{align}
	\hat {I}_{t-k}= \emph{H} I_{t-k}.
\end{align}
Here, $H$ represents the perspective transformation matrix between $I_{t-k}$ and $I_t$, $ \hat {I}_{t-k}$ denotes the motion-compensated previous $k$ frame. 

\subsubsection{Motion feature extraction.}
After we have obtained the motion-compensated previous $k$ frame, we can highlight the moving areas with absolute differences between the current frame and the motion-compensated previous $k$ frame. Frame difference is a simple but effective technique to find the change, it can capture the pixel-level change between images. If we assume the acquisition time between neighboring frames is in a short range, we can expect that those differences originate from obviously moving objects. In general, there are many different versions of frame difference, such as 2-frame difference, 3-frame difference, and 5-frame difference. In this paper, a modified 3-frame difference is used to extract motion features. We first calculate the frame difference between two adjacent frames. The frame difference is defined as follows:
\begin{align}
        E_{1,2} &= |I_t - \hat {I}_{t-k}|, \\
        E_{2,3} &= |I_t - \hat {I}_{t+k}|,
\end{align}
where $I_t$ is the gray value of the current frame, $ \hat {I}_{t-k}$ and $ \hat {I}_{t+k}$ are the gray value of the motion-compensated previous $k$ frame and next $k$ frame.

Next, based on extensive experiments, a threshold of 5 is applied on $E_{1,2}$ and $E_{2,3}$ to remove noise and highlight the silhouette of the potential moving objects. As a result, two binarized frame differences $D_{1,2}$ and $D_{2,3}$ are obtained. Finally, logical OR operation is applied on two binarized frame difference masks to enhance the motion feature $E_{t}$ as follows:
\begin{align}
        E_{t} &= D_{1,2} \cup D_{2,3}.
\end{align}

 Different from the classical three-frame difference method that uses logical AND operation to suppress noise and obtain a more intact outline, logical OR operation can enhance motion features of tiny moving objects by aggregating moving areas from multiple frames. A detailed comparison between different frame difference methods will be given in Section \ref{ablation}.

\subsubsection{Post-processing.}
Frame difference can segment the pixels belonging to moving objects, but there are also many noise, small holes, and disconnected blobs. Hence, morphological operation and connected component analysis are used to obtain more precise bounding boxes. Firstly, the morphological open and close operation with a structure element of size 3$\times$3 is used iteratively to eliminate isolated pixels and fill the holes. Then, connected component analysis is exploited to obtain the total number of pixels of each object, and the blob below 15 pixels is eliminated because such a small object usually belongs to interruption or alignment error. Finally, the minimum bounding rectangle is applied to mark the moving objects and obtain the final bounding boxes.

\subsection{Trajectory filtering}
We have obtained the candidate moving objects after the motion feature enhancement. However, there are still many false positives such as cars, pedestrians, swaying trees and image alignment errors. Especially for image alignment errors, the false positives generated by motion parallax are severe when MAV flies at low altitudes within urban areas. To remove these false positives, we first adopt multi-object tracking to generate a trajectory for each object. Then, based on the motion feature analysis, we can eliminate these false positives by trajectory classification. 

\subsubsection{Trajectory generation.}
To generate the trajectory for each candidate moving object, we adopt the classical multi-object tracking algorithm SORT\cite{sort}. The SORT uses a Kalman filter with a constant velocity motion model to predict the tracked objects' center, size, and aspect ratio in the current frame. Data association is viewed as a linear assignment problem that can be solved by the Hungarian algorithm \cite{hungarian}. If the intersection over union (IOU) between one detection and one tracking result is over 0.3, the detection is associated with that trajectory. Otherwise, a new tracking is activated. The max age of each trajectory is set to 3. It means that if one trajectory fails to be associated with any detection for successive three frames, the trajectory would be deleted. The min-hits are set to 1 to start a trajectory immediately after a successful detection.

\subsubsection{Trajectory classification.}
In the previous step, the trajectories of candidate moving objects have been generated with multi-object tracking. To remove trajectories belonging to image alignment errors and swaying trees, a trajectory classification algorithm based on motion analysis is proposed. In particular, the proposed trajectory classification model mainly considers the temporal and spatial features of candidate moving objects.

\textit{1) Temporal feature:} By observing the motion feature of moving objects and false positives, we found that the movement of MAVs is usually continuous and lasts for several frames. However, the movement of false positives such as swaying trees and image alignment errors usually lasts for a short time. Therefore, we assign an object with a high probability of being a moving object if it moves for a long time. In particular, we view the object with an age over 3 frames as a potential moving object.

\textit{2) Spatial feature:} On the other hand, we notice that the false positives usually move within a small range. As a comparison, real moving objects such as pedestrians, cars, and MAVs usually move over a wide range. Therefore, we give objects with large distances between consecutive frames with a high probability of being an MAV. In particular, the distance metric is defined as
\begin{align}
	D(b_i, b_j)=  \frac {{\rho(b_i, b_j)}} {d(b_i, b_j)},
\end{align}
where $b_i$ and $b_j$ are bounding boxes of the same object in the $i$-th frame and $j$-th frame, $\rho(b_i, b_j)$ represents the central Euclidean distance between $b_i$ and $b_j$, $d(b_i, b_j)$ represents the diagonal length of the smallest enclosing box of $b_i$ and $b_j$. 

\begin{figure}[h]
	\centering
	\subfloat[Results \textit{before} trajectory filtering.]{\includegraphics[width=0.98\linewidth]{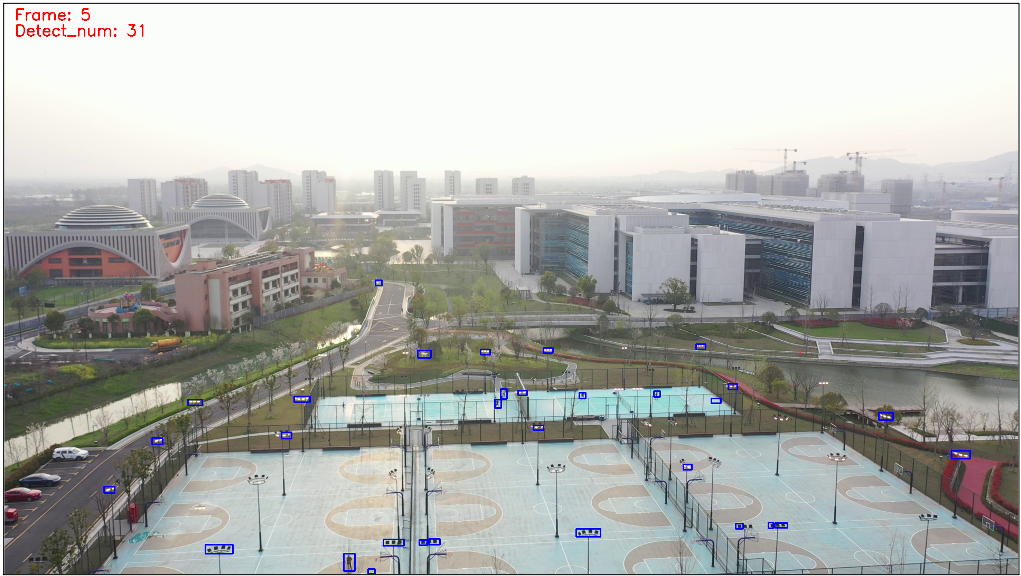}}\hspace{1mm}
	\subfloat[Results \textit{after} trajectory filtering.]{\includegraphics[width=0.98\linewidth]{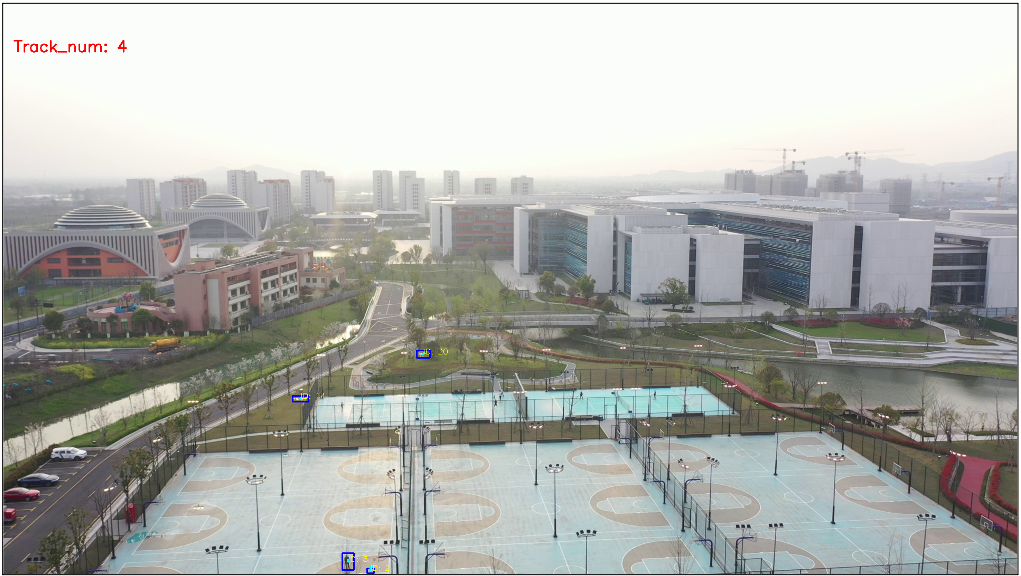}}
	\caption{An example of trajectory filtering. After the trajectory filtering, most of the bounding boxes generated by image alignment errors are removed.}
	\label{fig6}
\end{figure}

\textit{3) Trajectory Classification:} The details of the algorithm calculating the probability of each target being a potential moving target or a false positive are given in Algorithm 1. For each target, we define $S = \{s_1, s_2, . . . , s_N\}$ as the set of all the bounding boxes of the target over $N$ frames. We first align $s_{N-3}$ to the frame of $s_{N}$ by the previously mentioned perspective transformation. Then, we calculate the distance metric between $s_{N-3}$ and $s_N$ and classify the target as a moving object or a false positive directly based on the value of the distance metric. An example of trajectory filtering is demonstrated in Fig.~\ref{fig6}. As we can see, the image alignment errors such as lampposts and trees are mostly removed, and the true moving objects such as MAVs, cars, and pedestrians are reserved.

\renewcommand{\algorithmicrequire}{\textbf{Input:}}
\renewcommand{\algorithmicensure}{\textbf{Output:}}

\begin{algorithm}[h]
\caption{The trajectory classification algorithm}
\begin{algorithmic}[1]
\REQUIRE ~~\\
The bounding box of a potential moving target, $m_i$;\\
The set of stored $N$ bounding boxes of the potential moving target, $S = \{s_1, s_2, ..., s_N\}$;
\ENSURE ~~\\
The probability of a potential moving target, $P$;
\STATE DEFINITION $f(\cdot)$ a function counting the size of a set;
\STATE Put $m_i$ into $S$ at the last position;
\IF {$f(S) \ge 3$}
\STATE Align the $s_{N-3}$ to the frame of $s_N$;
\STATE Calculate $D(s_{N-3}, s_N)$;
\STATE $P =
\begin{cases}
	1, & { \text {if  } } D(s_{N-3}, s_N) \ge 1, \\
	{0,} &{ \text {if  } } D(s_{N-3}, s_N) < 1,
\end{cases};$\\
% \IF {$f(S) \ge 10$}
% \STATE Remove the first element $s_1$ from $S$;
% \ENDIF
\ELSE
\STATE {$P$ = 0;}
\ENDIF
%\RETURN $P$; 
\end{algorithmic}
\end{algorithm}

\subsection{Local fine detection}
After the trajectory filtering, the candidate moving objects still contain some false positives such as moving cars, pedestrians, flying birds, and a few image alignment errors. Moreover, the bounding boxes generated by motion feature enhancement are usually redundant with excessive background areas. Therefore, we apply a local appearance-based classifier (LAC) and a local appearance-based detector (LAD) to further eliminate false positives and obtain a refined bounding box.

\subsubsection{Local appearance-based classifier.}
The false positives such as moving cars, pedestrians, and flying birds usually have similar motion features but different appearance features with MAV, we can use an appearance-based classifier to classify them into MAV and clutter. Considering that the appearance-based classifier will be called frequently and most of the common CNN architectures have a very deep structure that requires large computation resources, a three-layer CNN network is used to extract the feature of MAV and classify candidate moving objects into MAV and clutter. We train the proposed CNN model with local images of candidate moving objects. The input images are resized to 32$\times$32 before being sent to the network. Each convolution layer uses ReLU activation and max pooling. The final layer uses a fully connected work with a softmax activation to classify the candidate moving objects into two classes.

\subsubsection{Local appearance-based detector.}
In the previous step, we obtained the candidate MAV targets with an appearance-based classifier. However, since some targets are too small to recognize and a shallow network can hardly extract enough high-level semantic information from the MAV target, the candidate MAV targets still contain some false positives that look similar to tiny MAVs. In addition, the bounding boxes generated through morphological operation are not accurate enough and have redundant enclosing regions. Therefore, an appearance-based detector is applied on the local region cropped from the neighboring area of the candidate MAV targets to obtain the precise position and final detection result.

In this paper, a 320$\times$320 area around the center of the candidate MAV target is cropped as the local search region. Then, YOLOv5s is applied to the cropped image for refined MAV detection. YOLO is a fast single-shot object detector based on convolutional neural networks. Our previous study \cite{2021Air} shows that YOLO can achieve a good balance between accuracy and speed. To improve its performance on small object detection, a small object detection layer is added to concatenate the shallow feature map with the deep feature map. The local detector is trained with cropped images from the training dataset and the target with a confidence over 0.5 is viewed as a successful detection result. %Up to now, we have successfully obtained the moving MAV. 

\section{Experiments}\label{experiment}

\subsection{Datasets}\label{dataset}
Most of the existing MAV datasets are either collected by stationary cameras such as \cite{2021DT-Benchmark, 2022Anti-UAV-DT} or have relatively large object sizes and simple backgrounds \cite{2021Fast}. As far as we know, there are few datasets specifically designed for small MAV detection under complex and non-planar scenes from an aerial view. To evaluate the performance of the proposed MGMD algorithm, we test our algorithm on our previously proposed ARD-MAV dataset\footnote{https://github.com/WestlakeIntelligentRobotics/Global-Local-MAV-Detection}. This dataset contains 60 video sequences and 106,665 frames. All the videos are taken by the cameras of DJI Mavic2 or DJI M300 flying at low and medium altitudes. The videos contain many real-world challenges such as complex backgrounds, non-planar scenes, occlusion, abrupt camera movement, fast-moving MAVs, and small MAVs. Each video is about one minute long with a 30 FPS frame rate and has a resolution of 1920$\times$1080. The object size ranges from 6$\times$3 to 136$\times$75. The average object size is only 0.02$\%$ of the image size. As far as we know, this is the smallest average object size among the existing MAV datasets. We use 50 videos for training and 10 videos for testing. In particular, the test videos are mainly composed of small MAVs and MAVs under complex backgrounds.

\subsection{Metrics and implementation details.}

\textbf{Metrics.} Following the related works\cite{2021Dogfight}, in this experiment, the performance evaluation is based on commonly used metrics such as precision, recall, F1-Score, and average precision (AP). Since the targets in this dataset are too small and a small deviation of the position of the bounding box will result in a large change of intersection over union (IOU), we set the IOU threshold between predictions and ground truth to 0.25. Therefore, detected targets matching with ground truth with IOU $>$ 0.25 are counted as true positives. 

\textbf{Implementation details.}
Our experiments are implemented on a computer with an Intel i7 processor, 32GB RAM, and an NVIDIA Geforce RTX 3070 GPU. For the training of the local fine detector, the input image size is set to 320$\times$320. We use the Adam optimizer with a momentum of 0.937 and an initial learning rate of 0.01. The training is started from publicly available pre-trained model weights of YOLOv5s on MS-COCO. For the appearance-based classifier, the Adam optimizer is applied with a learning rate of 0.001. We trained the model for 100 epochs with a batch size of 64.

\subsection{Comparison with state-of-the-art methods}\label{comparison}
We compare our proposed algorithm with recently published methods such as one-stage object detector, aerial object detector \cite{TPH-YOLOv5}, video object detector\cite{mega}, and aerial small object detector \cite{2021Dogfight}. In particular, to avoid the resolution loss introduced by the down-sampling method, YOLOv5s, TPH-YOLOv5s, and TPH-YOLOv5l use the 1536$\times$1536 input image size for inference rather than the default 640$\times$640. All compared methods are implemented based on official codes and fine-tuned using the pre-trained weights available with public codes.

The quantitative comparison of MGMD with the compared methods on the ARD-MAV dataset is shown in Table~\ref{tab1}. The experimental results demonstrate that our proposed algorithm outperforms the existing methods on various metrics. Specifically, our proposed method outperforms the best prior method by 19$\%$ on the AP metric. In addition, our method can run with 28 frames per second (FPS).

\begin{figure*}[t]
        \centering
	{\includegraphics[width=0.95\linewidth]{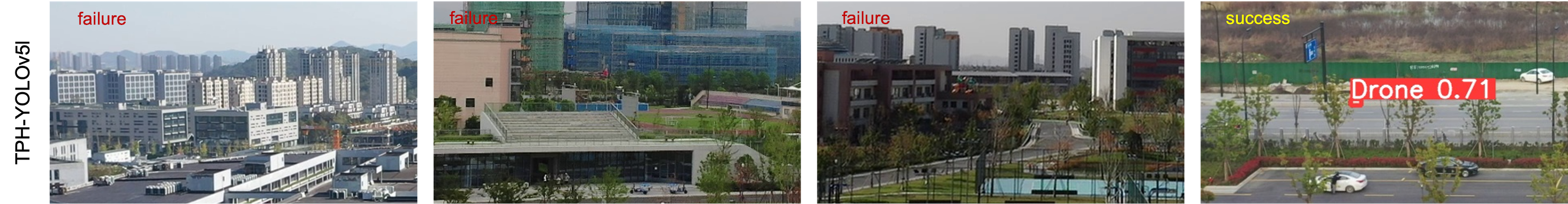}}\vspace{0.5mm}
        {\includegraphics[width=0.95\linewidth]{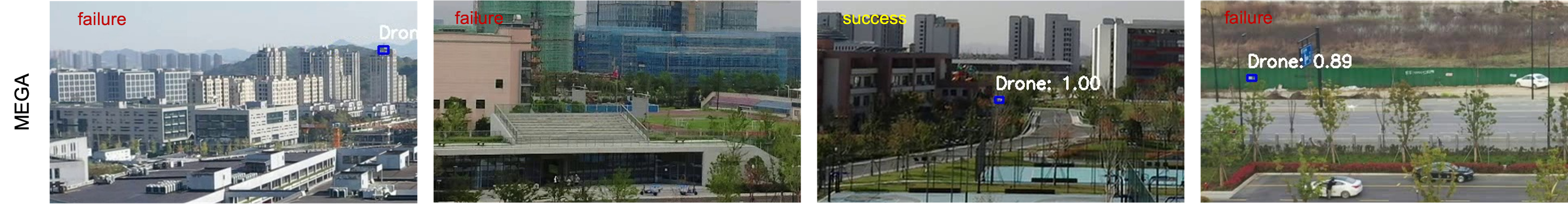}}\vspace{0.5mm}
	{\includegraphics[width=0.95\linewidth]{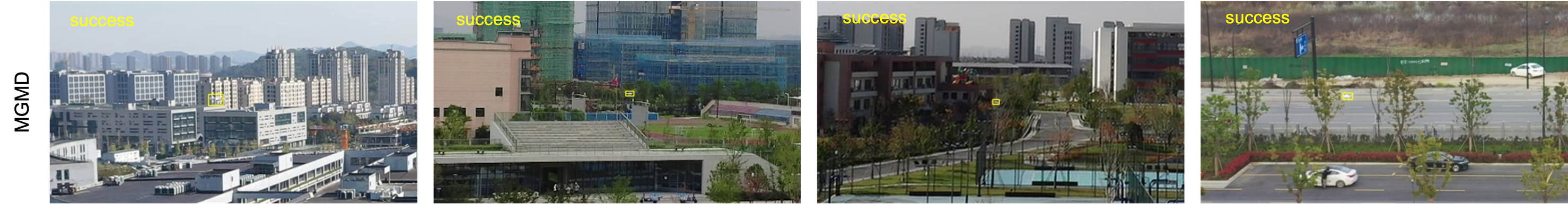}}
	\caption{The qualitative comparison between the detection results from TPH-YOLOv5l, MEGA, and MGMD. MGMD can detect the target MAV when the appearance features are unreliable under challenging conditions. \textbf{Yellow box} indicates the target detected by our proposed method. \textbf{Red box} indicates the target detected by TPH-YOLOv5l. \textbf{Blue box} indicates the target detected by MEGA. Because the image size is too large compared to the target's size, we only show the \textit{cropped regions} from the full-size image for a better view.}
	\label{fig_8}
\end{figure*}

The qualitative comparison of MGMD with the compared methods is shown in Fig.~\ref{fig_8}. As we can see, the proposed method can detect very tiny MAVs and MAVs under complex backgrounds. However, the compared method can hardly detect a target MAV under such challenging conditions. It is important to note that the small MAV here denotes the MAV with a size smaller than 10$\times$10 pixels, which is difficult even for humans to recognize in the image.

\begin{table}[h]
	% \begin{center}
            \centering
		\caption{Quantitative comparison of the MGMD with state-of-the-art methods on ARD-MAV dataset.}
		\label{tab1}
            \fontsize{8}{10.8}\selectfont%设置字体大小
		% \begin{tabular}{cccccc}
            \vspace{-8pt}
            \begin{tabularx}{\linewidth}{lXXXXr}
			\toprule %[1pt]
			Method & P & R & F1 & AP & FPS \\
			\midrule %[1pt]
                YOLOv5s & 0.86 & 0.27 & 0.41 & 0.26 & \textbf{82}\\
			TPH-YOLOv5s\cite{TPH-YOLOv5} & 0.81 & 0.30 & 0.44 & 0.34 & 35\\
                TPH-YOLOv5l\cite{TPH-YOLOv5} & \textbf{0.99} & 0.32 & 0.48 & 0.36 & 11\\
			Dogfight\cite{2021Dogfight} & 0.77 & 0.24 & 0.36 & 0.24 & 1\\
			MEGA\cite{mega} & 0.26 & 0.37 & 0.31 & 0.23 & 3\\
			MGMD & 0.84 & \textbf{0.59} & \textbf{0.69} & \textbf{0.55} & 28\\
			\bottomrule %[1pt]
		% \end{tabular}
            \end{tabularx}
	% \end{center}
\end{table}

\subsection{Ablation studies}\label{ablation}
To examine the effectiveness of different components in MGMD, we conduct extensive experiments to study how they contribute to the final performance.%The experimental results are shown in Table~\ref{tab2}.}

\textbf{Motion Feature Enhancement.} We have tested different frame difference methods such as 2-frame difference, and classical 3-frame difference to verify the effectiveness of our proposed motion feature enhancement module. The experimental results in Table~\ref{tab2} show that our proposed 3-frame difference with OR operation can effectively improve the recall while sacrificing a small amount of precision and inference speed.
% \captionsetup[table]{labelfont=bf,labelsep=newline,position=top,justification=raggedright}

\begin{table}[!h]
	% \begin{center}
            \centering
		\caption{Ablation study of different frame difference methods.}
		\label{tab2}
            \fontsize{8}{10.8}\selectfont%设置字体大小
            \vspace{-8pt}
		% \begin{tabular}{ccccc}
            \begin{tabularx}{\linewidth}{lXXXr}
			\toprule %[2pt]
			Method & P & R & F1 & FPS\\
			\midrule %[2pt]
                2-frame  & 0.82 & 0.45 & 0.58 & \textbf{43}\\
                3-frame + AND & \textbf{0.87} & 0.46 & 0.61 & 33\\
                3-frame + OR & 0.84 & \textbf{0.59} & \textbf{0.69} & 28\\
			\bottomrule %[2pt]
		% \end{tabular}
            \end{tabularx}
	% \end{center}
\end{table}

\textbf{Trajectory Filtering.} This paper mainly uses the TF module to remove false positives such as swaying trees and image alignment errors in non-planar scenes. As shown in Table~\ref{tab3}, when the TF module is introduced, the precision and inference speed are improved in most cases. However, the TF module sometimes reduces the recall as it mistakenly removes the hovering or slowly moving MAVs.

\textbf{Local Appearance-based Classifier.} As shown in the first and third row of Table~\ref{tab3}, the LAC module significantly improves detection accuracy and inference speed. This is because the LAC module can effectively remove interruptions with similar motion features but looks distinctly different from MAV. As a result, the workload of the LAD module is greatly reduced and the detection accuracy and inference speed are improved.

\textbf{Local Appearance-based Detector.} As shown in the second and third row of Table~\ref{tab3}, the LAD plays an important role by promoting the precision metric. Since the LAC works on the segmented images generated by motion feature enhancement and is trained with a shallow network, it can hardly discriminate false positives that look similar to tiny MAVs and have a coarse bounding box. However, the LAD is trained with ground-truth labels, it has a bounding box refinement module and better semantic representation towards MAVs. Therefore, it can significantly improve the detection accuracy by eliminating false positives that look similar to tiny MAVs.

\begin{table}[ht]
	% \begin{center}
            \centering
		\caption{Ablation study of different components in MGMD.}
		\label{tab3}
            \fontsize{8}{10.8}\selectfont%设置字体大小
            \vspace{-8pt}
		\begin{tabularx}{\linewidth}{lXXXr}
			\toprule %[2pt]
                % \hline
			Method & P & R & F1 & FPS\\
			\midrule %[2pt]
                MFE + LAD  & 0.60 & 0.58 & 0.59 & 16\\
                MFE + LAC  & 0.40 & 0.39 & 0.39 & \textbf{34} \\
                MFE + LAD + LAC & 0.79 & \textbf{0.63} & \textbf{0.70} & 23\\
                MFE + TF + LAD & 0.79 & 0.58 & 0.67 & 22\\
                MFE + TF + LAC & 0.45 & 0.37 & 0.41 & 32\\
                MFE + TF + LAC + LAD & \textbf{0.84} & 0.59 & 0.69 & 28\\
                % \hline
			\bottomrule %[2pt]
		\end{tabularx}
	% \end{center}
\end{table}

\subsection{Discussion}
\begin{figure}[h]
	\centering
	\subfloat[The input image.]{\includegraphics[width=0.45\linewidth]{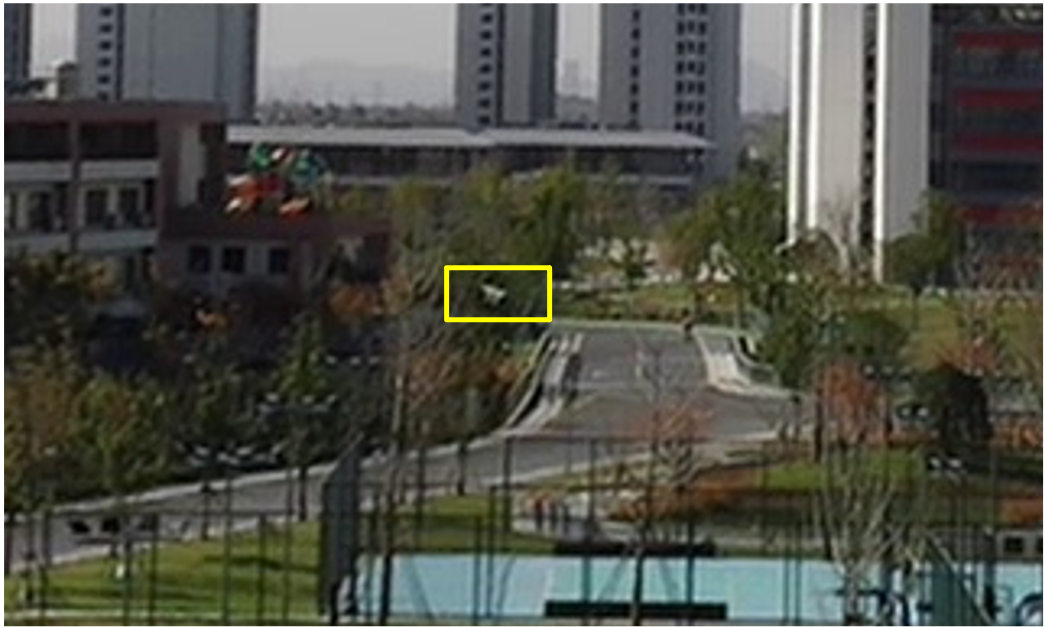}}\hspace{1mm}
        \subfloat[Heatmap of YOLOv5.]{\includegraphics[width=0.45\linewidth]{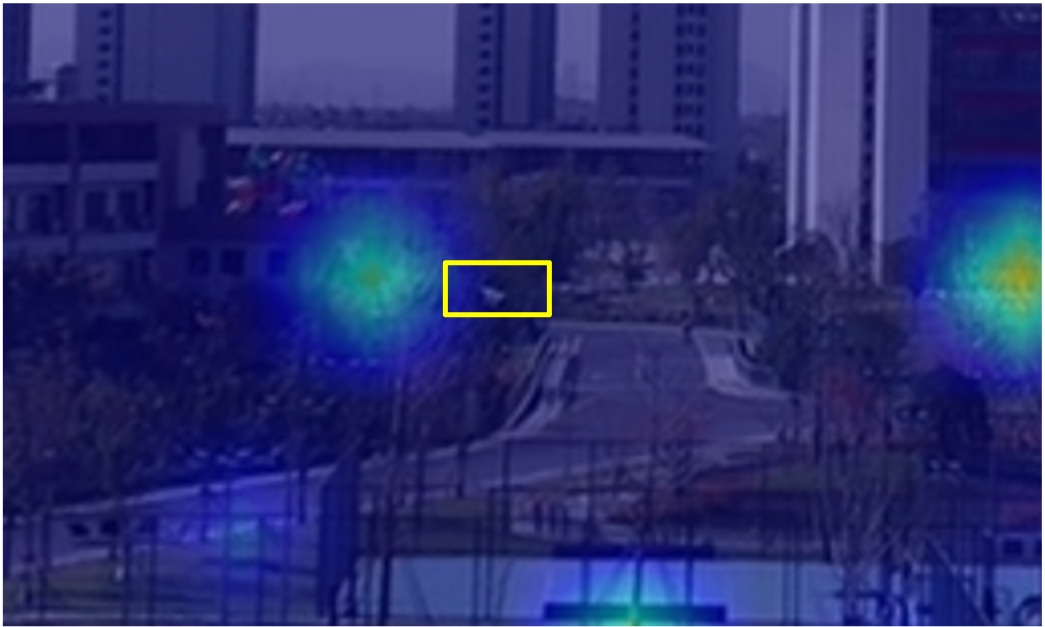}}\vspace{1mm}
	\subfloat[The motion boundary extracted by optical flow gradient\cite{2021Dogfight}.]{\includegraphics[width=0.45\linewidth]{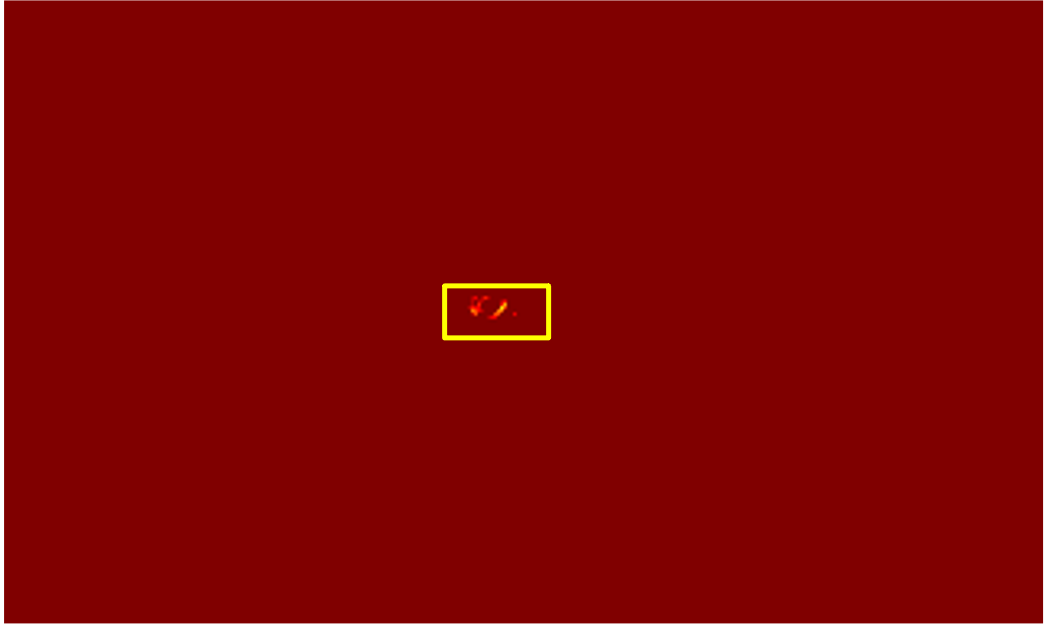}}\hspace{1mm}
	\subfloat[The motion feature extracted by our method.]{\includegraphics[width=0.45\linewidth]{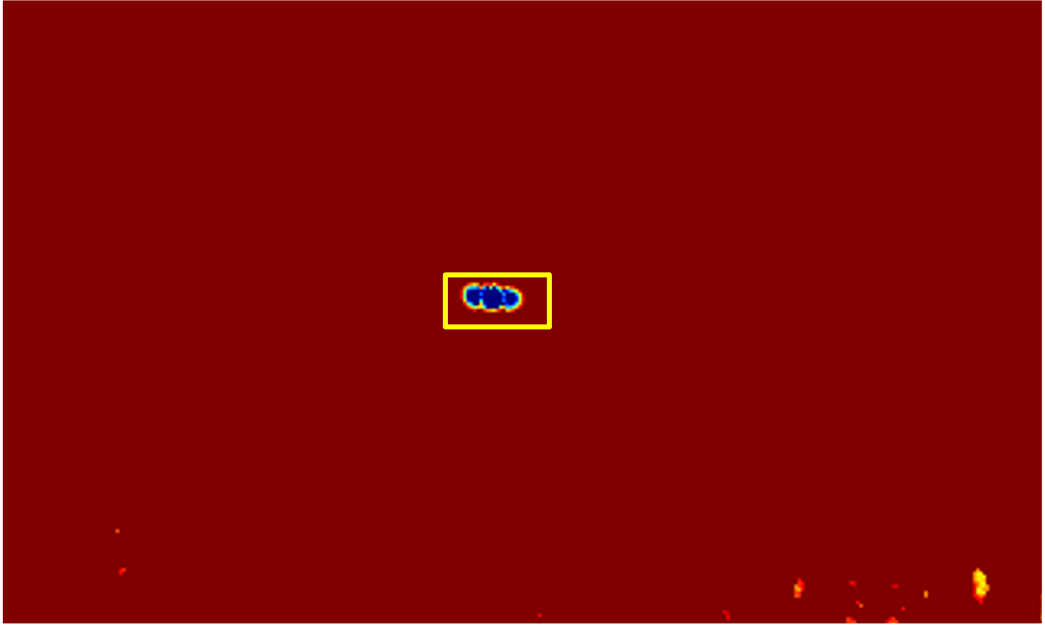}}
	\caption{A comparison of the appearance feature and motion feature generated by different methods.} 
	\label{fig9}
\end{figure}

%Our method mainly focuses on detecting small MAVs under complex and non-planar scenes. 
Experimental results demonstrate that our proposed method outperforms the state-of-the-art methods on various metrics. We attribute it to several factors. First, MGMD adopts a motion feature enhancement module to extract motion information and can detect the target MAV when the target is extremely small or the background is complex. However, appearance-based detectors such as YOLOv5 and TPH-YOLOv5 fail to capture the appearance feature in these scenarios. In addition, the existing motion-based methods such as \cite{2021Dogfight} also have difficulties under such challenging conditions. Some examples are shown in Fig.~\ref{fig9}. 
Second, MGMD uses a local fine detector which works on a local search region cropped from the full-size image. This local search region greatly retains the valuable appearance information of small targets and eliminates the influence of other interruptions that look similar. Therefore, it achieves better performance on the precision metric. Third, the multi-object tracking method makes good use of the temporal feature of MAV and eliminates the interruptions that occasionally emerge.

\section{Conclusion}\label{conclusion}
In this paper, we propose a simple yet effective algorithm for small MAV detection in complex and non-planar scenes. Our main contribution is to extract the motion features of extremely small MAVs using the proposed motion feature enhancement module, then utilize a trajectory-based classifier and an appearance-based classifier to eliminate false positives generated by motion parallax and other moving objects. In addition, a modified YOLOv5s is applied to the cropped regions to achieve refined detection results. Ablation experiments show the effectiveness of each module in our method. Together, our proposed algorithm could achieve an excellent performance of 0.55 AP on the ARD-MAV dataset which contains various challenging conditions, and achieve a running speed of 28 FPS.

Although the proposed method has made some progress, it still faces the problem of false detections when detecting small targets with high similarity to MAVs such as tiny birds and cars. In addition, our proposed method mainly focuses on moving targets, the hovering and slowly moving MAVs are usually ignored. In the future, we will try to extend our algorithm to both moving and hovering MAVs. Moreover, an end-to-end network that can learn motion cues and appearance features within a unified network is necessary to be designed to simplify the training process and reduce the empirical parameters.

\section*{Acknowledgments}
This should be a simple paragraph before the References to thank those individuals and institutions who have supported your work on this article.

\bibliography{reference_guo} % if no referece is cited before, there will be an error
\bibliographystyle{ieeetr}

\vfill

\end{document}